\documentclass[a4paper]{article}

\usepackage[hyperref]{INTERSPEECH2021}

\usepackage{times}
\usepackage{latexsym}

\usepackage{microtype}
\usepackage{hyperref}

\usepackage{graphicx}
\usepackage{xspace}
\usepackage{paralist}
\usepackage{booktabs}
\usepackage{multirow}
\usepackage{amsmath}
\usepackage{amsfonts}
\usepackage{amssymb}
\usepackage{pifont}
\usepackage{mathrsfs}
\usepackage{algorithm,algpseudocode}
\usepackage{mdwlist}
\usepackage{enumitem}
\usepackage{color}
\usepackage{dsfont}
\usepackage{amsthm}
\usepackage{bm}
\usepackage{subfig}

\newcommand{\loss}{\ensuremath{\mathcal{L}}}

\newcommand{\expect}{\ensuremath{\mathbb{E}}}

\newcommand{\notes}[1]{}

\theoremstyle{definition}

\theoremstyle{plain}

\newcommand{\vecs}{\ensuremath{\mathbf{s}}\xspace}

\newcommand{\ith}[1]{\ensuremath{i^{{th}}}}

\newcount\permx
\newcount\permy
\def\permdot#1#2{
\permx=#1 \advance\permx by-1
\permy=#2 \advance\permy by-1
\psframe[fillcolor=black, fillstyle=solid]
(\permx,\permy)(#1, #2)
}

\newcommand\union{\cup}

\newcommand{\boxnum}[1]{{\setlength{\fboxsep}{1pt}\raisebox{1pt}{\hspace{1pt}\fbox{\tiny #1}\hspace{1pt}}}}
\newcommand{\ind}[1]{\ensuremath{_{\kern-0.5pt\boxnum{#1}}}}

\newcommand{\vecx}{\mathbf{x}\xspace}
\newcommand{\vecy}{\mathbf{y}\xspace}
\newcommand{\vece}{\ensuremath{\bm{e}}\xspace}

\newcommand{\smallnt}[1]{\ensuremath{_{\mbox{\tiny PP}}}\xspace}

\newcommand{\pseudocode}{Algorithm}
\floatname{algorithm}{\pseudocode}

\iffalse

\else

\fi

\newcommand{\method}{XSTNet\xspace}

\title{End-to-end Speech Translation via Cross-modal Progressive Training}
\name{Rong Ye, Mingxuan Wang, Lei Li}
\address{ByteDance AI Lab, Shanghai, China}
\email{\texttt{\{yerong,wangmingxuan.89,lileilab\}@bytedance.com}}

\begin{document}

\maketitle

\begin{abstract}
End-to-end speech translation models have become a new trend in research due to their potential of reducing error propagation. 
However, these models still suffer from the challenge of data scarcity. 
How to effectively use unlabeled or other parallel corpora from machine translation is promising but still an open problem.
In this paper, we propose \textbf{Cross} \textbf{S}peech-\textbf{T}ext \textbf{Net}work (\textbf{\method}), an end-to-end model for speech-to-text translation. \method takes both speech and text as input and outputs both transcription and translation text. 
The model benefits from its three key design aspects:
a self-supervised pre-trained sub-network as the audio encoder, a multi-task training objective to exploit additional parallel bilingual text, and a progressive training procedure. We evaluate the performance of \method and baselines on the MuST-C En-X and LibriSpeech En-Fr datasets. In particular, \method achieves state-of-the-art results on all language directions with an average BLEU of 28.8, outperforming the previous best method by 3.2 BLEU.
Code, models, cases, and more detailed analysis are available at \url{https://github.com/ReneeYe/XSTNet}.
\end{abstract}

\section{Introduction}
\label{sec:intro}
Speech-to-text translation (ST) has found increasing applications. It takes speech audio signals as input and outputs text translations in the target language.
Recent work on ST has focused on unified end-to-end neural models with the aim to supersede pipeline approaches combining automatic speech recognition (ASR) and machine translation (MT). 
However, training end-to-end ST models is challenging - there is limited parallel \textit{speech-text} data.
For example, there are only a few hundreds hours for English-German in MuST-C corpus.
Existing approaches to this problem can be grouped into two categories:
\begin{inparaenum}[\it a)]
    \item multi-task supervision with the \textit{speech-transcript-translation} triple data~\cite{indurthi2020data,tang2021general};
    \item pre-training with external large-scale MT parallel text data~\cite{dong2021consecutive, wang2020curriculum,bansal2019pre, zheng2021fused}.
\end{inparaenum}
We notice that the triple data can decompose into three sub-tasks with parallel supervision, ST, ASR, and MT. This motivates us to design a multi-task model.

In this paper, we designed \textbf{Cross} \textbf{S}peech-\textbf{T}ext \textbf{Net}work (\textbf{\method}) for end-to-end ST to joint train ST, ASR and MT tasks.
\method supports either audio or text input and shares a Transformer~\cite{vaswani2017attention} module.
To bridge the gap between the audio and text modality, we use a self-supervised trained Wav2vec2.0 representation of the audio~\cite{baevski2020wav2vec}, which provides a more compressed and contextual representation than the log-Mel filter bank feature.
Furthermore, our method is able to incorporate external large-scale MT data.
Finally, to support the training of \method, we carefully devise the progressive multi-task learning strategy, a multi-stage procedure following the popular pre-training and fine-tuning paradigm.

Despite the model's simplicity, the experimental results on MuST-C and Augment LibriSpeech datasets improved by a big margin (+3.2 BLEU) against the previous SOTA method.

\section{Related Work}
\label{sec:related}
\noindent\textbf{End-to-end ST}~
\cite{berard2016listen} gave the first proof of the potential for end-to-end ST without using the intermediate transcription. 
And recent Seq2Seq models have received impressive results~\cite{kano2017structured,berard2018end,inaguma2020espnet, zhao2021neurst}. 
Techniques, such as pre-training~\cite{wang2020curriculum,bansal2019pre,alinejad2020effectively,dong2021consecutive}, multi-task learning~\cite{le2020dual,tang2021general}, self-training~\cite{pino2020self} and meta-learning~\cite{indurthi2020data} have further improved the performance of end-to-end ST models.
Very recently, \cite{zheng2021fused} proposed a Fused Acoustic and Text Masked Language Model to pre-train and improved ST by fine-tuning.

\noindent\textbf{Self-supervised Pretraining}~
This work is partially motivated by the recent success of self-supervised contrastive learning for speech~\cite{schneider2019wav2vec, baevski2020vq, baevski2020wav2vec}.
These representations have been shown to be effective in low-resource ASR~\cite{baevski2020wav2vec}, ST~\cite{wu2020self, nguyen2020investigating} and multi-lingual ST~\cite{tran2020cross}.
Recently, many audio-related tasks have demonstrated the feasibility and outstanding performance by using the wav2vec2.0 representation~\cite{wang2021unispeech, yi2021applying}. 
Maybe the most related work is~\cite{tran2020cross}. However, our work focuses more on the training strategy, while theirs focused on incorporating two pre-trained modules to strengthen the multi-lingual ST.

\section{Proposed Method: \method}
\label{sec:approach}

\subsection{Problem Formulation}


The speech translation corpus contains \textit{speech-transcript-translation} triples 
$\mathcal{D} = \{(\vecs, \vecx, \vecy)\}$, where $\vecs=(s_1,...,s_{|\vecs|})$ is the input sequence of the audio wave (or the acoustic features), 
$\vecx=(x_1,...,x_{|\vecx|})$ is the transcript from the source language, 
and $\vecy=(y_1,...,y_{|\vecy|})$ represents the corresponding translation in the target language.
Despite the fact that transcripts are provided during training, the end-to-end speech translation models directly produce the translation $vecy$ without producing the transcript $vecx$ as an intermediate output. As a result, understanding how to leverage ancillary transcript supervision and make the most of the triple-supervised dataset is critical.
The triple-supervised dataset can be pairwise combined into three parallel-supervised sub-datasets, $\mathcal{D}_{\text{ASR}}=\{(\vecs, \vecx)\}$, $\mathcal{D}_{\text{ST}}=\{(\vecs, \vecy)\}$ and $\mathcal{D}_{\text{MT}}=\{(\vecx, \vecy)\}$ , which solve ASR, end-to-end ST and MT respectively.

We also introduce external MT dataset $\mathcal{D}_{\text{MT-ext}}=\{(\vecx',\vecy')\}$.
The amount of external MT corpus is much larger than the ST corpus, i.e. $|\mathcal{D}_{\text{MT-ext}}| \gg |\mathcal{D}|$. 


\begin{figure}[t]
    \setlength{\abovecaptionskip}{-0.cm}
    \setlength{\belowcaptionskip}{-0.5cm}
    \centering
    \includegraphics[scale=0.38]{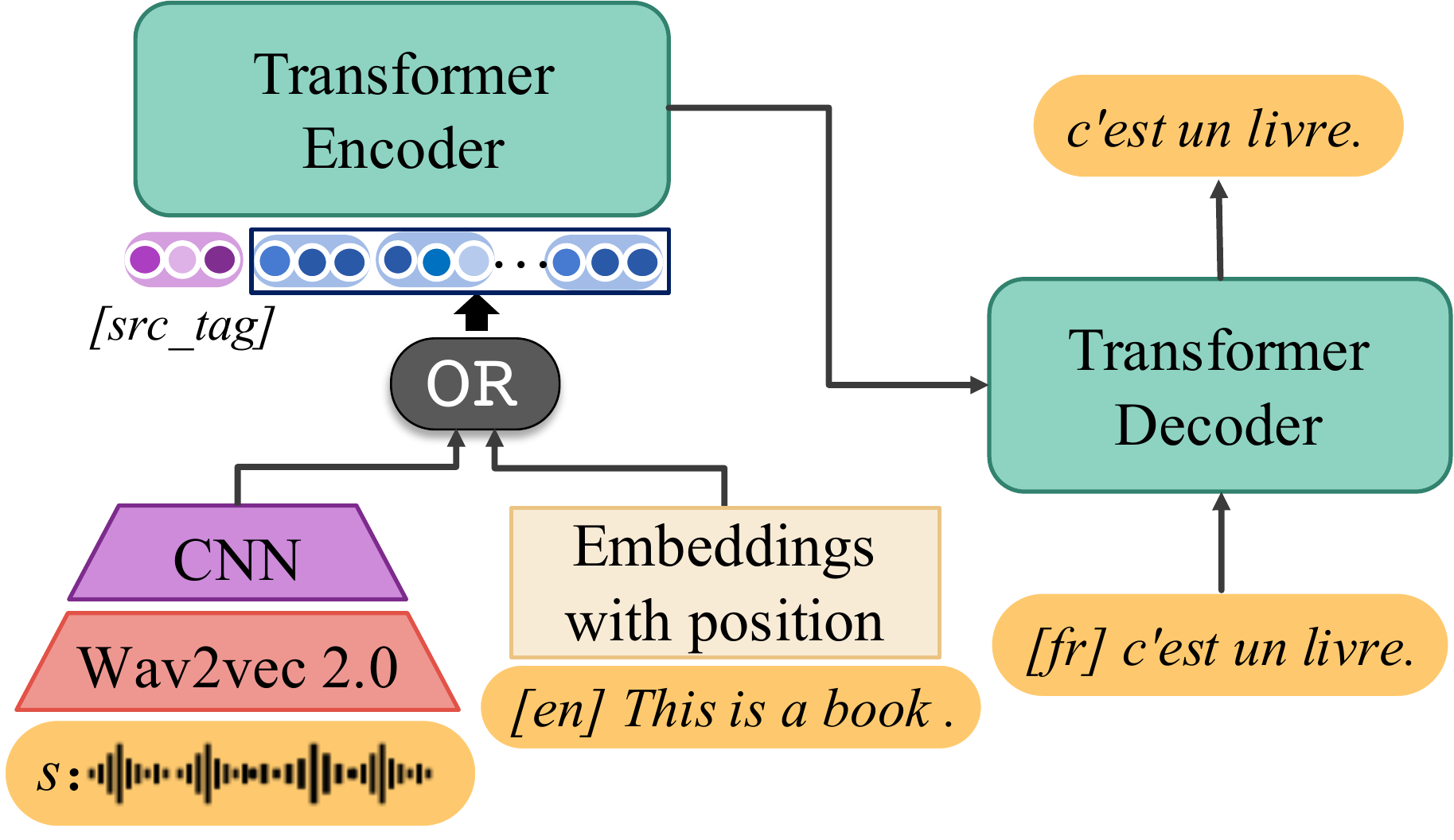}
    \caption{The model structure of \method. It accepts either audio or text input. For the audio input, we deploy wav2vec2.0 followed by two convolution layers to get the audio embedding. The Transformer encoder and decoder are shared for both modalities.}
    \label{fig:model}
\end{figure}

\subsection{Speech Encoder}
\label{sec:model:w2v}
The first part of \method is Wav2vec2.0 sub-network to process speech data in waveform.
\noindent\textbf{Wav2vec2.0}~\cite{baevski2020wav2vec} is a model to learn the contextualized speech representation from unlabelled audio data.
It consists of a multi-layer convolutional feature encoder and a Transformer-based context encoder. 
The multi-layer convolutional encoder takes the raw audio signal as input and outputs the latent speech representation, which is then used by the Transformer encoder to output the contextual representation.
The self-supervised pre-trained contextual audio representation $\ensuremath{\bm{c}}\xspace =[c_1, ...c_T]$ offers a good initialization for the model's audio presentation.
In the experiment, we use the raw 16-bit 16kHz mono-channel audio as the audio input and follow a base configuration of wav2vec2.0, trained on audio data from Librispeech~\cite{panayotov2015librispeech} but without fine-tuning\footnote{\url{https://dl.fbaipublicfiles.com/fairseq/wav2vec/wav2vec_small.pt}}.

\noindent\textbf{Convolution layers}~
The sequence of Wav2vec2.0 output for a voice utterance typically has a length of a few hundred, larger than that of a sentence. 
To further match the lengths of the audio representation and text sequences, we add two additional layers of 2-stride 1-dimensional convolutional layers with GELU activation after Wav2vec2.0, reducing the time dimension by a factor of 4.
Hence, we have $\vece_s\xspace =\text{CNN}(\ensuremath{\bm{c}})$, $\vece_s \in \mathds{R}^{d \times T/4}$, where $d$ is the same as the hidden size of Transformer.
We set the kernel size of CNN 5, and the hidden size $d=512$.

\subsection{Encoder-Decoder with Modality and Language Indicators}
\label{sec:model:transf}
Since \method supports both speech and text as input, we add an additional lookup table for text tokens to map the embeddings.
We adopt the standard Transformer-base model~\cite{vaswani2017attention} to accomplish the specific generation tasks, ST, ASR, and MT. 
We use pre-layer normalization for stable training.

We use different indicators \texttt{[src\_tag]} to distinguish the three tasks and audio/text inputs.
The input embedding $\vece$ (either audio or text) is fed into the Transformer encoder.
Specifically, 
\begin{itemize}[topsep=0pt, itemsep=0pt, leftmargin=10pt]
    \item For the audio input, we add extra \texttt{[audio]} token with embedding $\vece_{[audio]} \in \mathds{R}^{d}$, and the embedding of the audio $\vece \in \mathds{R}^{d \times (T/4+1)}$ is the concatenation of $\vece_{[audio]}$ and $\vece_s$ in terms of the sequence length.
    
    \item For the text input, we put the language id symbol before the sentence. For example, the embedding of English sentence ``\texttt{This is a book.}" is the embedding of ``\texttt{[en] This is a book.}". 
\end{itemize}

When decoding, the language id symbol serves as the initial token to predict the output text. For example, if the audio input for sentence ``\texttt{This is a book.}" is in English, to do ASR, 
we use \texttt{[en]} as the BOS and decode ``\texttt{[en] This is a book.}", while to translate into French,
we use \texttt{[fr]} as the BOS and decode ``\texttt{[fr] c'est un livre.}"

\subsection{Progressive Multi-task Training}
\label{sec:method:train}
The progressive multi-task training strategy consists of two stages, large-scale MT pre-training with external parallel text and multi-task fine-tuning.

\noindent\textbf{Large-scale MT Pre-training}~
We first pre-train the transformer encoder-decoder module using external MT data $\mathcal{D}_{\text{MT}}'$. 
\begin{equation}
    \loss(\theta) = - \expect_{\vecx, \vecy \in \mathcal{D}_{\text{MT-ext}}} \log P(\vecy | \vecx;\theta)
    \label{eq:pretrain}
\end{equation}
, where $\theta$ is the model parameters. 
Experimental results show that the MT pre-training provide a good warm-up for the shared transformer module.

\noindent\textbf{Multi-task Fine-tuning}~
During the fine-tuning, we combine external MT, ST, ASR, and MT parallel data from the in-domain speech translation dataset and jointly optimize the negative log-likelihood loss.

\begin{equation}
    \loss(\theta) = - \expect_{\vecx, \vecy \in \mathcal{D} \union \mathcal{D}_{\text{MT-ext}}} \log P(\vecy | \vecx; \theta) 
    \label{eq:mlt-tune}
\end{equation}
,where $\mathcal{D} = \mathcal{D}_{\text{ST}} \union \mathcal{D}_{\text{ASR}} \union \mathcal{D}_{\text{MT}}$ is the union set of all the parallel subsets (the same notation hereinafter).

The overall training process is shown in Algorithm~\ref{alg:train}.
It is \textit{\textbf{progressive}}, because the external MT data, used in the pre-training stage, is continuously used in the fine-tuning stage. In Section~\ref{sec:051_progressive_train}, we will show that the training procedure largely influences the translation performance, and the progressive multi-task training process is the most effective.
In the experiment, we use Adam optimizer~\cite{kingma2015adam} with learning rate = $2\times 10^{-4}$ and warm-up 25k steps.

\begin{algorithm}[t]
    \caption{Progressive Multi-task Training for \method}
    \label{alg:train}
    \begin{algorithmic}[1] 
        \State \textbf{Input}: Tasks T=$\{\text{ST}, \text{ASR}, \text{MT}, \text{MT-ext}\}$
        \State Initialize the speech module parameters in $\theta$ using wav2vec2.0-base, and the rest at random. 
        \State MT pre-training: optimize Eq~(\ref{eq:pretrain}) on $\mathcal{D}_{\text{MT-ext}}$.
        \While{not converged}
            \State \textbf{Step 1}: random select a task $\tau$ from T.
            \State \textbf{Step 2}: sample a batch of $(\vecx,\vecy)$ from $\mathcal{D}_{\tau}$, and optimize the cross entropy loss defined in Eq~(\ref{eq:mlt-tune}).
        \EndWhile
    \end{algorithmic}
\end{algorithm}

\section{Experiments}
\label{sec:exps}

\subsection{Datasets}
\label{sec:exp:data}

\noindent\textbf{ST datasets}~
We conduct experiments on \textbf{MuST-C}~\cite{digangi2019must} and \textbf{Augmented LibriSpeech} En-Fr (LibriTrans)~\cite{kocabiyikoglu2018augmenting} datasets. MuST-C contains Engish speech to 8 languages: German (De), Spanish (Es), French (Fr), Italian (It), Dutch (Nl), Portuguese (Pt), Romanian (Ro), and Russian (Ru). We evaluate the BLEU scores on \texttt{tst-COMMON} of MuST-C and the test set of LibriTrans.

\noindent\textbf{MT datasets}~
We use external WMT machine translation datasets\footnote{\url{ https://www.statmt.org/wmt16/translation-task.html}}~\cite{bojar2016findings} for En-De/Es/Fr/Ro/Ru directions, and OPUS100\footnote{\url{http://opus.nlpl.eu/opus-100.php}}~\cite{zhang2020improving} for En-It/Nl/Pt directions. We also introduce OpenSubtitle\footnote{\url{https://opus.nlpl.eu/OpenSubtitles-v2018.php}}~\cite{lison2016opensubtitles2016} for En-De.


\begin{table*}[!th]
    \setlength{\abovecaptionskip}{-0.3cm}
    \setlength{\belowcaptionskip}{-0.cm}
    \centering
    \resizebox{\textwidth}{!}{
    \begin{tabular}{l|c|c||cccccccc|c}
        \toprule
        
        \textbf{Models} & \textbf{External Data} & \textbf{Pre-train Tasks} &
        \textbf{De} & \textbf{Es} & \textbf{Fr} & \textbf{It} &	
        \textbf{Nl} & \textbf{Pt} & \textbf{Ro} & \textbf{Ru} & \textbf{Avg.}\\
        \midrule

        Transformer ST~\cite{zhao2021neurst}& \texttimes & ASR &
            22.8 & 27.4 & 33.3  & 22.9 & 27.2 & 28.7 & 22.2 & 15.1 & 24.9 \\
        
        AFS~\cite{zhang2020adaptive} & \texttimes & \texttimes & 
            22.4 & 26.9 & 31.6 & 23.0 & 24.9 &26.3 & 21.0 & 14.7 & 23.9 \\

        Dual-Decoder Transf.~\cite{le2020dual} & \texttimes & \texttimes &
        23.6 &28.1 & 33.5 & 24.2 & 27.6 & 30.0 & 22.9 & 15.2 & 25.6 \\

        
        Tang et al.~\cite{tang2021general}  & MT & ASR, MT & 
            23.9 & 28.6 & 33.1 & - & - & - & - & - & -\\
        
        FAT-ST (Big)~\cite{zheng2021fused} & ASR, MT, mono-data$^\dagger$ & FAT-MLM &
            25.5 & 30.8 & - & - & 30.1 & - & - & - & - \\
            
        \midrule
       
        W-Transf. & audio-only* & SSL* & 
            23.6 & 28.4 & 34.6 & 24.0 & 29.0 & 29.6 & 22.4 & 14.4 & 25.7 \\

        \textbf{\method (Base)} & audio-only* & SSL* & 
            25.5 & 29.6 & 36.0 & 25.5 &	30.0 &31.3 & 25.1 &	16.9 & 27.5 \\
        
        \textbf{\method (Expand)} & MT, audio-only* & SSL*, MT &
            \textbf{27.8}$^\mathsection$ & \textbf{30.8} & \textbf{38.0} & \textbf{26.4} & 
            \textbf{31.2} & \textbf{32.4} & \textbf{25.7} & \textbf{18.5} & \textbf{28.8} \\

        \bottomrule
    \end{tabular}
    }
    \caption{Performance (case-sensitive detokenized BLEU) on MuST-C test sets. 
    $^\dagger$: ``Mono-data'' means audio-only data from Librispeech, Libri-Light, and text-only data from Europarl/Wiki Text; 
    *: ``Audio-only'' data from LibriSpeech is used in the pre-training of wav2vec2.0-base module, and ``SSL'' means the self-supervised learning from unlabeled audio data.
    $^\mathsection$ uses OpenSubtitles as external MT data.
    }
    \label{tab:vs_baseline_all}
\end{table*}

\begin{table}[t!]
    \setlength{\abovecaptionskip}{-0.7cm}
    \setlength{\belowcaptionskip}{-0.cm}
    \centering
    \small
    \resizebox{0.42\textwidth}{!}{
    \begin{tabular}{l|cc}
        \toprule
        \textbf{Models} & \textbf{Uncased-tok} & \textbf{Cased-detok}\\
        \midrule
        
        Transformer ST~\cite{zhao2021neurst}& 18.7 & 16.3 \\
        AFS~\cite{zhang2020adaptive} & 18.6 & 17.2 \\
        Curriculum~\cite{wang2020curriculum}  & 18.0 & - \\
        LUT~\cite{dong2021listen} & 18.3 & - \\
        STAST~\cite{liu2020bridging} & 18.7 & - \\
        \midrule
        W-Transf. & 14.6 & 13.0\\
        \textbf{\method (Base)}  & 21.0 & 18.8 \\
        \textbf{\method (Expand)}  & \textbf{21.5} & \textbf{19.5} \\
        \bottomrule
    \end{tabular}}
    
    \caption{Performance (case-insensitive tokenized BLEU and case-sensitive detokenized BLEU) on LibriTrans En-Fr test set.}

    \label{tab:libritrans}
\end{table}

\subsection{Experimental Setups}
\label{sec:exp:setups}


We jointly tokenize the bilingual text (En and X) using subword units with a vocabulary size of 10k, learned from SentencePiece\footnote{\url{https://github.com/google/sentencepiece}}~\cite{kudo2018sentencepiece}.
The model configurations have been stated in Section~\ref{sec:model:w2v} and~\ref{sec:model:transf}.
We save the checkpoint with the best BLEU on dev-set and average the last 10 checkpoints. We use a beam size of 10 for decoding. 
We report the case-sensitive detokenized BLEU using sacreBLEU\footnote{\url{https://github.com/mjpost/sacrebleu}}~\cite{post2018call} for MuST-C dataset. We also list both case-insensitive tokenized (Uncased-tok) BLEU and case-sensitive detokenized (Cased-detok) BLEU for LibriTrans for a fair comparison.

\subsection{Main Results}
We compared our method with other strong end-to-end baseline models, including:
Transformer ST~\cite{zhao2021neurst}, 
AFS~\cite{zhang2020adaptive},
Dual-decoder Transformer~\cite{le2020dual}, 
STAST~\cite{liu2020bridging}, 
Tang et al.~\cite{tang2021general},
Curriculum pretrain~\cite{wang2020curriculum}, LUT~\cite{dong2021listen}, 
and FAT-ST~\cite{zheng2021fused}.
These baselines all take the 80-channel log Mel-filter bank feature (fbank80) as the audio inputs. 
We also implement wav2vec2+transformer model (abbreviated \textbf{W-Transf.}) with the same configuration as \method.
\textbf{\method (Base)} is only trained based on MuST-C by optimizing $\loss = - \expect_{\vecx, \vecy \in \mathcal{D}} \log P(\vecy | \vecx)$.
\textbf{\method (Expand)} uses external WMT data and follows the progressive multi-task training method described in Section~\ref{sec:method:train}. 

Table~\ref{tab:vs_baseline_all} and Table~\ref{tab:libritrans} respectively show the BLEU scores of MuST-C and LibriTrans datasets.
Compared to the current speech transformer, which acts as a solid baseline for ST, our method achieves a remarkable +3.9 BLEU gain on average.
We attribute the gain to the following factors: 

\noindent\textbf{Wav2vec2.0 vs.~Fbank}~
Comparing the models using the fbank80 feature with the ones applying wav2vec2, we find that even the simplest structure (W-Transf.) without ASR pre-training outperforms the Transformer ST baseline.
This indicates the potential of wav2vec2.0 representation.

\noindent\textbf{Multi-task vs.~ST-only}~
Comparing \method (Base) and W-Transf., we find that the translation performance improved by around +2 BLEU in all directions.
This demonstrates that our model can make the most of the triple-supervised data by applying the multi-task paradigm.

\noindent\textbf{Additional MT data}~
Since \method accepts text input, it is easy to introduce additional MT parallel data.
\method (Expand) attributes an average of +1.3 BLEU improvement against \method-Base from the additional MT data and the training procedure described in Section~\ref{sec:method:train}.

\subsection{Results on Auxiliary MT and ASR Tasks}
\method can perform MT and ASR tasks provided different input modalities and target language indicators.
The performance of auxiliary MT and ASR (measured in WER) tasks are shown in Table~\ref{tab:mt_result} and Table~\ref{tab:asr_result}.
Our \method with progressive training is better than other training strategies in both tasks.
Comparing Row ``I'' and ``IV'' in Table~\ref{tab:mt_result}, it is worth noting that adding speech-to-text data improves MT performance even further. We attributed the improvement to the information from the speech.

\begin{table}[ht]
    \setlength{\abovecaptionskip}{-0.8cm}
    \setlength{\belowcaptionskip}{-0.cm}
    \centering
    \small
    \resizebox{0.93\columnwidth}{!}{
    \begin{tabular}{l|l|ccc}
        \toprule
        \textbf{Pretrain} & \textbf{Finetune} & \textbf{En-De} & \textbf{En-Fr} & \textbf{En-Ru} \\
        \midrule
        - & $\mathcal{D}_{\text{MT}} \union \mathcal{D}_{\text{MT-ext}}$ & 31.65 & 43.46 & 21.06 \\
        \midrule
        - & $\mathcal{D}$ & 30.74 & 42.91 & 19.54 \\
        $\mathcal{D}_{\text{MT-ext}}$ & $\mathcal{D}$ & 32.82 & 44.82 & 21.29 \\
        $\mathcal{D}_{\text{MT-ext}}$ & $\mathcal{D}\union \mathcal{D}_{\text{MT-ext}}$ & \textbf{33.20} & \textbf{45.25} & \textbf{21.51} \\
        \bottomrule
    \end{tabular}}
    \caption{Comparisons of the auxiliary MT tasks among different multi-task training strategies. The model in the first row is trained based on the external MT data $\mathcal{D}_{\text{MT-ext}}$ and the \textit{transcript-translation} parallel data in MuST-C.}
    \label{tab:mt_result}
\end{table}

\begin{table}[ht]
    \setlength{\abovecaptionskip}{-0.7cm}
    \setlength{\belowcaptionskip}{-0.cm}
    \centering
    \small
    \resizebox{0.91\columnwidth}{!}{
    \begin{tabular}{l|l|ccc}
        \toprule
        \textbf{Pretrain} & \textbf{Finetune} & 
        \textbf{En-De} & \textbf{En-Fr} & \textbf{En-Ru} \\
        
        \midrule
        - & $\mathcal{D}_{\text{ASR}}$ & 12.99 & 12.27 & 12.05 \\
        \midrule
        - & $\mathcal{D}$ & 11.08 & 10.74 & 10.94 \\
        $\mathcal{D}_{\text{MT-ext}}$ & $\mathcal{D}$ & 10.98 & 10.44 & 10.90 \\
        $\mathcal{D}_{\text{MT-ext}}$ & $\mathcal{D}\union \mathcal{D}_{\text{MT-ext}}$ & \textbf{10.80} & \textbf{10.38} & \textbf{10.86} \\
        \bottomrule
        
    \end{tabular}}
    \caption{Comparisons of the auxiliary ASR tasks, measured by word error rate (WER\textdownarrow) versus the reference English transcript.}
    \label{tab:asr_result}
\end{table}

\begin{table*}[th]
    \setlength{\abovecaptionskip}{-0.5cm}
    \setlength{\belowcaptionskip}{-0.cm}
    \centering
    \small
    \normalsize
    \resizebox{0.78\textwidth}{!}{
    \begin{tabular}{c|c|l|l|cccc}
        \toprule
         & \textbf{\#Exp.} & \textbf{Pretrain} & \textbf{Finetune} & \textbf{En-De} & \textbf{En-Fr} & \textbf{En-Ru} & \textbf{Avg.} \\
         \midrule
         
         \multirow{3}*{\rotatebox{90}{\small \textbf{MLT}}}& 
         I & $\mathcal{D}_{\text{MT-ext}}$ & $\mathcal{D} \union \mathcal{D}_{\text{MT-ext}}$ &
                \textbf{27.12} & \textbf{38.01} & \textbf{18.36} & \textbf{27.8}\\
            
         & II & $\mathcal{D}_{\text{MT-ext}}$ & $\mathcal{D}$ & 
                26.99 & 37.37 & 18.03 & 27.5 (-0.3)\\
        
         & III & - & $\mathcal{D} \union \mathcal{D}_{\text{MT-ext}}$ & 
                26.28 & 36.15 & 17.33 & 26.6 (-1.2)\\        
        
         \midrule
         
         \multirow{3}*{\rotatebox{90}{\small \textbf{ST-only}}} &  
         
         IV & $\mathcal{D}_{\text{MT-ext}} \rightarrow  \mathcal{D}_{\text{ASR}} \union \mathcal{D}_{\text{MT}} \union \mathcal{D}_{\text{MT-ext}} $ & $\mathcal{D}_{\text{ST}}$ & 
                26.96 & 35.42 & 17.87 & 26.6 (-1.2)\\
                
         & V & $\mathcal{D}_{\text{MT-ext}} \rightarrow \mathcal{D}_{\text{ASR}} \union \mathcal{D}_{\text{MT}}$ & $\mathcal{D}_{\text{ST}}$ & 
                25.86 & 34.44 & 16.94 & 25.6 (-2.2)\\
                
         & VI & $\mathcal{D}_{\text{MT-ext}} \union \mathcal{D}_{\text{MT}} \rightarrow \mathcal{D}_{\text{ASR}}$ & $\mathcal{D}_{\text{ST}}$ & 
                24.32 & 35.30 & 17.50 & 25.7 (-2.1)\\
        
        \bottomrule
    \end{tabular}}
    \caption{The ablation study results on the pre-train and fine-tuning strategies, where $\mathcal{D} = \mathcal{D}_{\text{ST}} \union \mathcal{D}_{\text{ASR}} \union \mathcal{D}_{\text{MT}}$, and Exp.I uses the training strategy described in Section~\ref{sec:method:train}.
    Experiments are preformed on MuST-C En-De, En-Fr and En-Ru datasets.}
    \label{tab:ablation}
\end{table*}

\subsection{Comparsion with Cascaded Baselines}
We also compare \method to the cascaded models, i.e. Transformer ASR \textrightarrow Transformer MT in Table~\ref{tab:vs_cascade}.
We also implemented a strong cascaded model whose ASR part is trained based on W-Transf., and the MT part is Transformer-base model trained based on $\mathcal{D}_{\text{MT}} \union \mathcal{D}_{\text{MT-ext}}$.
The results show that \method-expand translates better, meaning that \method can avoid the error propagation problem that plagues the cascaded models.

\begin{table}[ht]
    \setlength{\abovecaptionskip}{-0.75cm}
    \setlength{\belowcaptionskip}{-0.cm}
    \centering
    \small
    \resizebox{0.95\columnwidth}{!}{
    \begin{tabular}{c|l|ccc}
        \toprule
        & Models & \textbf{En-De} & \textbf{En-Fr} & \textbf{En-Ru}\\
        \midrule
        \multirow{2}*{\small \textbf{Cascaded}} &
        Espnet\cite{inaguma2020espnet} & 23.6 & 33.8 &  16.4 \\
        & Our implement* & 25.2 & 34.9 & 17.0 \\
        \midrule
        \small \textbf{End-to-end}
        & \method & 
        \textbf{27.1} & \textbf{38.0} & \textbf{18.4} \\
        \bottomrule
    \end{tabular}
    }
    \caption{\method versus the cascaded models on MuST-C En-De, En-Fr and En-Ru test sets. \textbf{Our implement*} is a strong cascaded model composed of W-Transf. as ASR and Transformer-base trained on $\mathcal{D}_{\text{MT}} \union \mathcal{D}_{\text{MT-ext}}$ as MT.}
    \label{tab:vs_cascade}
\end{table}

\section{Analysis}
\label{sec:analysis}

\subsection{The Influence of Training Procedure}
\label{sec:051_progressive_train}
In this section, we investigate the impact of the training strategy.
We perform six groups of experiments with different pre-train and fine-tune strategies, categorizing into two groups according to the different fine-tuning tasks - \textbf{multi-task} fine-tuning (Exp.~I, II and III) and \textbf{ST-only} fine-tuning (Exp.~IV, V and VI).
The multi-task fine-tuning incorporates ST, ASR and MT tasks using MuST-C dataset $\mathcal{D}(=\mathcal{D}_{\text{ST}} \union \mathcal{D}_{\text{ASR}} \union \mathcal{D}_{\text{MT}})$ and optional $\mathcal{D}_{\text{MT-ext}}$.
In the ST-only fine-tuning group, we perform a two-stage pre-training. 
For example, Exp.~IV first pre-trains with external MT data, then pre-trains with ASR and all the MT parallel (both MuST-C and WMT) data, following the idea of ``\textbf{\textit{progressive}}".
Table~\ref{tab:ablation} illustrates the detailed training process and their BLEU scores, with the following empirical conclusions.

\noindent\textbf{MT pretraining is effective.}~
Comparing Exp.I and III, we find that canceling the pre-training using external WMT reduces the average performance by 1.2 BLEU.
Furthermore, the results of Exp.~III are inferior to those of Exp.~II.
These findings suggest that MT pre-training provides a strong initialization for the model, leading it to perform better.

\noindent\textbf{Don't stop training the data in the previous stage.}~
An interesting discovery is that data used in the previous training stage can also be helpful in the subsequent training stage. In other words, \textbf{progressive training works.}
Concretely, we can see from Exp. I vs. III and Exp. IV vs. V that the BLEUs decrease as we stop using the WMT data at the fine-tuning period.



\noindent\textbf{Multi-task fine-tuning is preferred.}~
Comparing the BLEU results in the multi-task and ST-only groups, the models trained from multi-task perform better than the models trained by ST-only tasks.

\subsection{Convergence Analysis}

\begin{figure}[ht]
    \setlength{\abovecaptionskip}{-0.01cm}
    \setlength{\belowcaptionskip}{-0.5cm}
    \centering
    \includegraphics[scale=0.28]{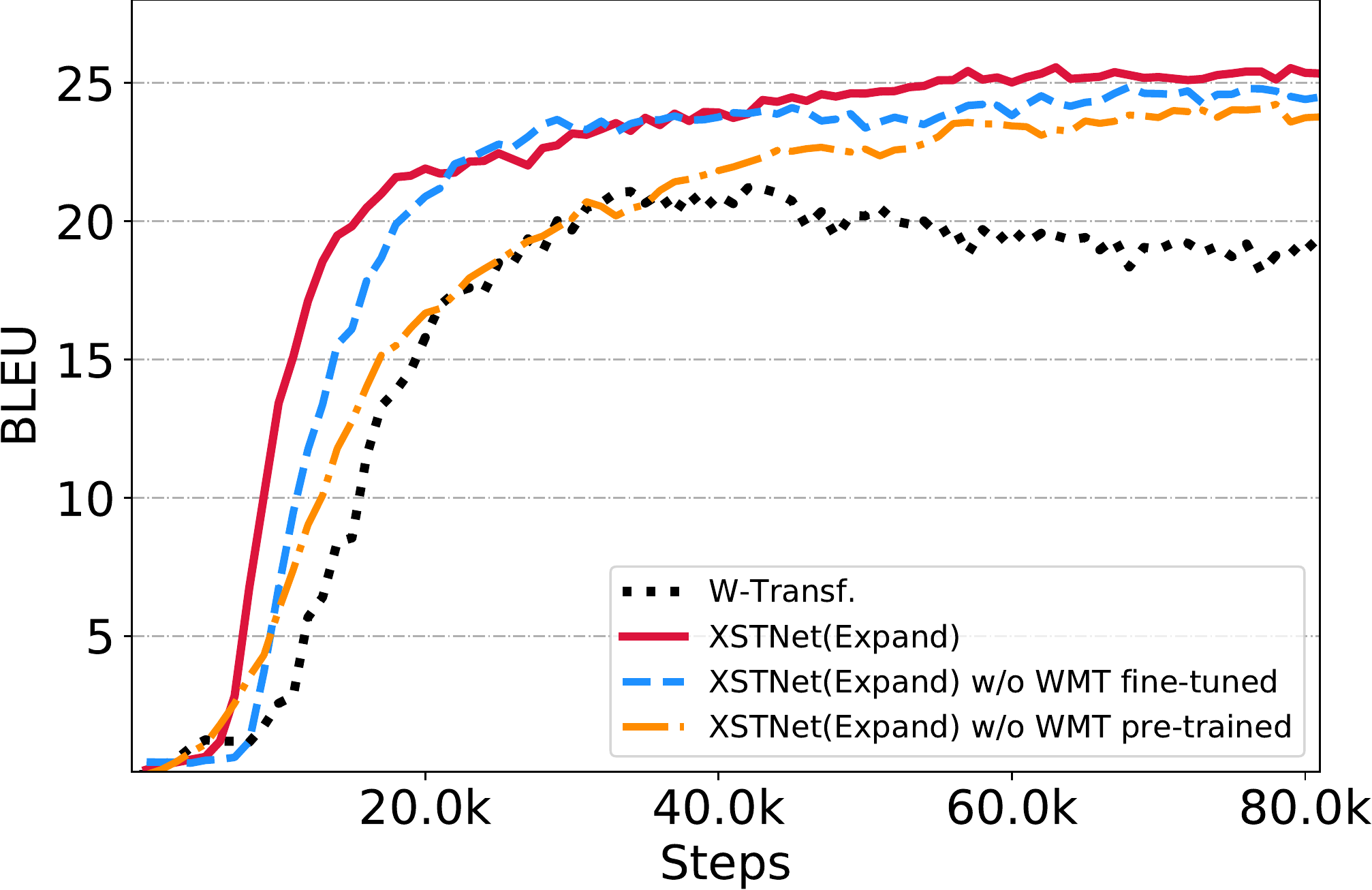}
    \caption{The evolution of BLEU on MuST-C En-De dev set.
    ``\method (Expand) w/o WMT pre-trained'' is trained directly on $\mathcal{D} \union \mathcal{D}_{\text{MT-ext}}$.
    ``\method (Expand) w/o WMT fine-tuned'' means fine-tuning only on $\mathcal{D}$, i.e. not progressively.
    The actual training has been performed until full convergence.
    }
    \label{fig:converge}
\end{figure}

Figure~\ref{fig:converge} depicts the evolution of the BLEU score over time steps for different training methods on the MuST-C En-De dev-set.

\noindent\textbf{Progressive multi-task training converges faster.}~
With WMT pre-trained parameters for the Transformer module, 
\method-Expand with progressive multi-task training (red line) is found to converge faster than the model without pre-training.

\noindent\textbf{Multi-task training generalizes better.}~
Due to the small data scale,
the model is tend to overfit the training set if it is trained only based on audio-translation parallel data.
However, by adopting a multi-task framework, the model is more robust and has a better generalization ability.

\subsection{Influence of Additional MT Data}
\begin{figure}[ht]
    \setlength{\abovecaptionskip}{-0.0cm}
    \setlength{\belowcaptionskip}{-0.6cm}
    \centering
    \includegraphics[scale=0.28]{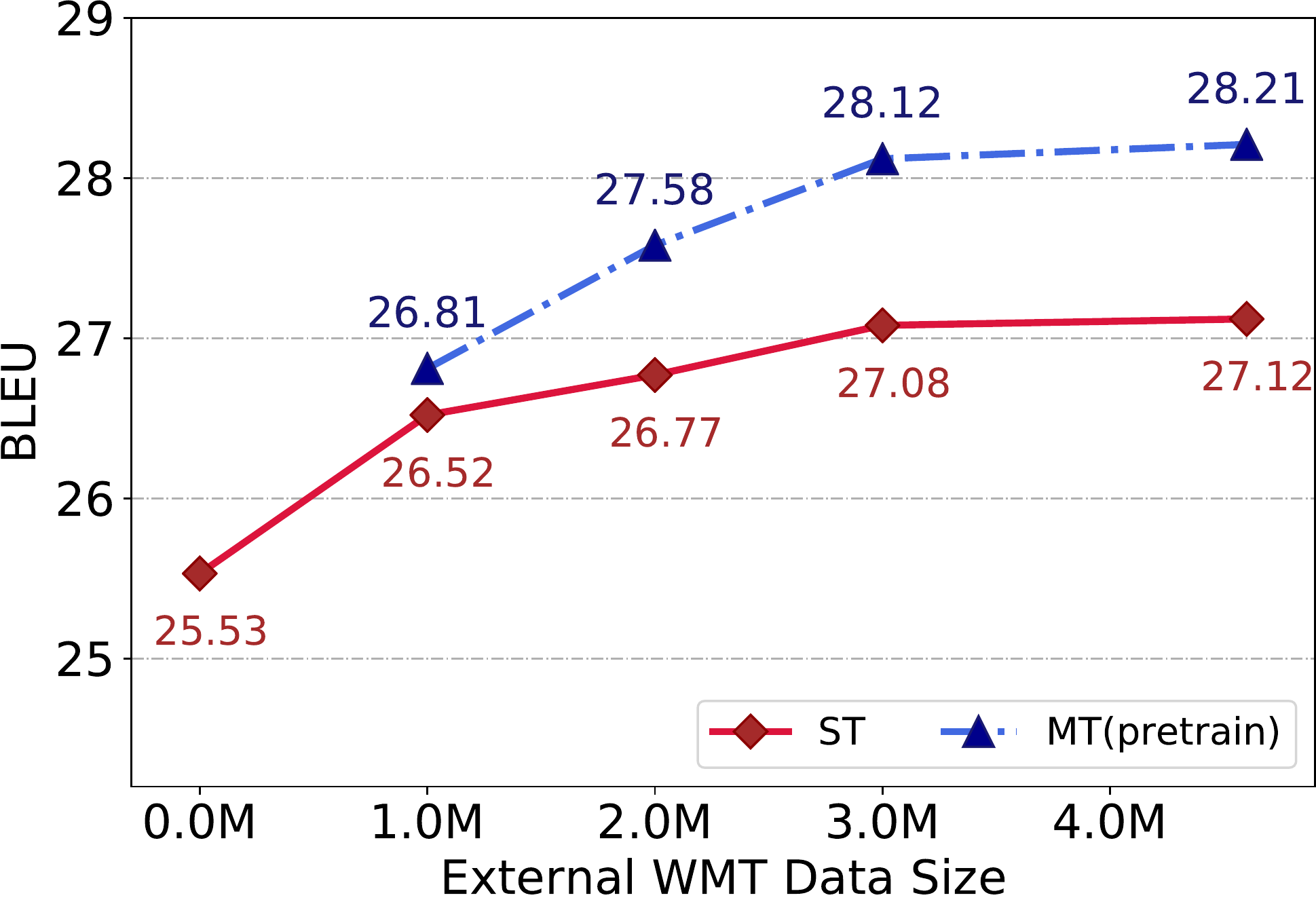}
    \caption{BLEU curve for MuST-C En-De against the amount of external MT data used. The red curve is the evolution of the ST BLEU. The blue curve is the evolution of the MT BLEU pre-trained based on WMT data.}
    \label{fig:WMT_vs_BLEU}
\end{figure}

We gradually increase the dataset size to assess the impact of the amount of external WMT data and test BLEU scores (Figure~\ref{fig:WMT_vs_BLEU}).
There are increases in the pretrained MT BLEU and ST BLEU as the increase of the external MT data size increases.
Substituting WMT with larger OpenSubtitles dataset (Table~\ref{tab:WMT_vs_opensubtitles}), the ST BLEU score of MuST-C En-De is even higher, achieving \textbf{27.8}.

\begin{table}[ht]
    \setlength{\abovecaptionskip}{-0.6cm}
    \setlength{\belowcaptionskip}{-0.cm}
    \centering
    \small
    \resizebox{0.86\columnwidth}{!}{
    \begin{tabular}{cc|cc}
         \toprule
        \textbf{Corpora} & \textbf{Size} & \textbf{MT}(Pretrain) & \textbf{ST} \\ 
        \midrule
        WMT & 4.6M & 28.2 & 27.1 \\
        OpenSubtitles & 18M & 28.5 & \textbf{27.8} \\
        \bottomrule
    \end{tabular}}
    \caption{BLEU scores for different external MT dataset (WMT v.s. OpenSubtitles)}
    \label{tab:WMT_vs_opensubtitles}
\end{table}

\section{Conclusion}
\label{sec:conclusion}
We propose Cross Speech-Text Network (\method), an extremely concise model which can accept bi-modal inputs and jointly train ST, ASR and MT tasks.
We also devise progressive multi-task training algorithm for the model.
As compared to the SOTA models, \method can achieve a significant improvement on the speech-to-text translation task.

\clearpage

\bibliographystyle{IEEEtran}
\bibliography{main}

\end{document}